# Hierarchical Attention for Sparse Volumetric Anomaly Detection in Subclinical Keratoconus

Lynn Kandakji, William Woof, Nikolas Pontikos

**Abstract**—The detection of weak, spatially distributed anomalies in volumetric medical imaging remains challenging due to the difficulty of integrating subtle signals across non-adjacent regions. This study presents a controlled comparison of sixteen architectures spanning convolutional, hybrid, and transformer families for subclinical keratoconus detection from three-dimensional anterior segment optical coherence tomography (AS-OCT). The results demonstrate that hierarchical architectures achieve 21-23% higher sensitivity and specificity, particularly in the difficult subclinical regime, outperforming both convolutional neural networks (CNNs) and global-attention Vision Transformer (ViT) baselines. Mechanistic analyses indicate that this advantage arises from spatial scale alignment: hierarchical windowing produces effective receptive fields matched to the intermediate extent of subclinical abnormalities, avoiding the excessive locality observed in convolutional models and the diffuse integration characteristic of pure global attention. Attention-distance measurements show that subclinical cases require longer spatial integration than healthy or overtly pathological volumes, with hierarchical models exhibiting lower variance and more anatomically coherent focus. Representational similarity further indicates that hierarchical attention learns a distinct feature space that balances local structure sensitivity with flexible long-range interactions. Auxiliary age and sex prediction tasks demonstrate moderately high cross-task consistency, supporting the generalizability of these inductive principles. The findings provide design guidance for volumetric anomaly detection and highlight hierarchical attention as a principled approach for early pathological change analysis in medical imaging.

**Index Terms**—Anomaly detection, optical coherence tomography, deep learning, hierarchical attention, risk stratification, medical image analysis, sparse volumetric data, 3D networks, volumetric transformers

## I. INTRODUCTION

The precise and efficient detection of weak, spatially sparse anomalies within three-dimensional (3D) medical imaging data presents a significant technical challenge for modern pattern recognition systems. Such anomalies, exemplified by conditions like subclinical keratoconus in anterior segment optical coherence tomography (AS-OCT), present as subtle, spatially dispersed deviations spread across multiple anatomical planes, requiring models that can integrate weak volumetric cues while suppressing measurement noise. This arises in diverse physical systems, including CT/MRI tumor microlesions, early organ-level pathology, and non-medical 3D inspection tasks, motivating a general study of architectural inductive biases for volumetric anomaly detection.

The optimal inductive bias for sparse volumetric anomaly detection is still unclear. Convolutional neural networks (CNNs) impose strong locality constraints through limited receptive fields; vision transformers (ViTs) enable unconstrained global attention at the cost of heavy data requirements [1, 2], and hierarchical transformers occupy an intermediate design space [3, 4]. While these families show distinct trade-offs on natural images [5], their relative performance on volumetric medical tasks with limited training data and spatially distributed signals has not been systematically characterized. Existing comparisons report accuracy but do not analyze why models succeed or fail, nor how their spatial integration mechanisms align with the distributed, low-contrast signatures of early disease. Medical volumes form a natural proving ground for these architectural questions: early disease signals are subtle and spatially extended [6, 7], and dataset sizes are comparatively small, making performance heavily dependent on inductive bias. The near-isotropic structure of CT, MRI, and OCT volumes further distinguishes medical imaging from natural-image benchmarks, where CNN locality and ViT globality behave differently. In volumetric data, CNNs struggle to integrate non-adjacent structure, while ViTs often diffuse attention across irrelevant regions.

Subclinical keratoconus (SKC) detection from 3D AS-OCT offers a rigorous testbed for these challenges. The

- The authors are all with the University College London Institute of Ophthalmology and Moorfields Eye Hospital, London, EC1V 9EL UK (e-mail: smgxlk0@ucl.ac.uk; w.woof@ucl.ac.uk; n.pontikos@ucl.ac.uk).

The work of L. Kandakji is supported by Moorfields Eye Charity under Grant GR001147 and a Doctoral Scholarship from Amazon Web Services. The work of W. Woof and N. Pontikos is supported by the National Institute for Health Research under Grant AI_AWARD02488, Sight Research UK (grant no. TRN004), and Medical Research Foundation and Moorfields Eye Charity (grant no. MRF-JF-EH-23-122). *(Corresponding author: Lynn Kandakji.)*

This work involved human subjects in its research. Approval of all ethical and experimental procedures and protocols was granted by the Institutional Review Board and the Ethics Committee of the UK Health Research Authority (reference 22/PR/0249). The study protocol was reviewed and approved by the Clinical Audit Assessment Committee of Moorfields Eye Hospital National Health Service Foundation Trust (reference CA17/CED/03). All research adhered to the tenets of the Declaration of Helsinki.

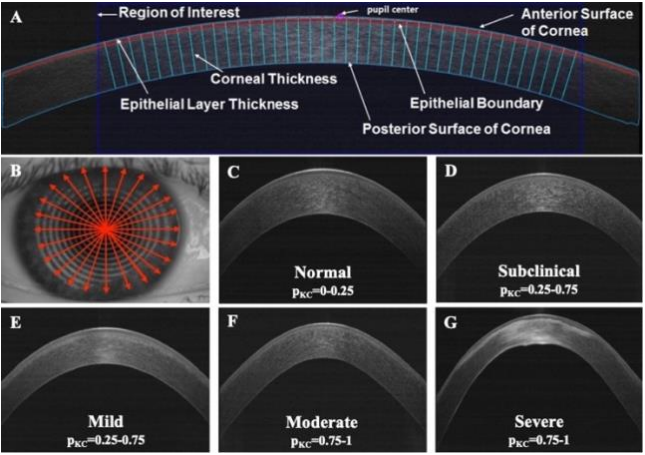

Fig. 1. Anterior segment optical coherence tomography scan geometry and representative cross-sections illustrating the subtle morphological progression with increasing keratoconus probability ($p_{KC}$). (A) Region of interest for anomaly detection. (B) Radial 360° scan pattern, comprised of 24 B-scan slices used to acquire each volume. (C–G) Example B-scans spanning the spectrum from normal to severe keratoconus, with corresponding keratoconus probability ranges $p_{KC}$. Keratoconus is characterized by progressive corneal thinning and anterior bulging, yet the earliest (subclinical) stage (D) presents only minute, spatially diffuse irregularities in curvature and epithelial/stromal thickness, changes that remain visually subtle even to trained clinicians.

earliest structural signatures are extremely subtle (Fig. 1), consisting of small, distributed shape irregularities and depth-dependent variations that required integration of weak volumetric cues [8, 9]. Prior models achieve high sensitivity for manifest disease but substantially lower performance in the subclinical regime [10]. The problem also naturally admits a continuous-risk formulation [11], where disease severity is modeled as a probability score derived from tomographic indices via Gaussian mixture modeling, avoiding hard categorical boundaries that introduce label noise for progressive diseases. Models are trained to predict a single image-level risk score from the entire volume, requiring spatial integration of distributed pathological cues without localization supervision. This weak-supervision regime tests whether architectural inductive biases alone can guide models to attend to diagnostically relevant regions.

In this work, we systematically compare 16 architectures spanning convolutional, hybrid, and transformer families in matched 2D and 3D configurations. Beyond performance comparison, we examine why architectures succeed or fail using effective receptive fields analysis, attention-distance distributions, and representational similarity via centered kernel alignment (CKA) [12]. The objective of this investigation is threefold. First, to establish whether volumetric models consistently outperform slice-based approaches across architectural families. Second, to determine whether hierarchical attention mechanisms provide systematic advantages over purely local CNNs and purely global ViTs for sparse volumetric anomaly detection. Third, to identify the architectural properties that govern performance on tasks requiring integration of weak, spatially distributed signals. Our main contributions can be summarized as follows:

1) A unified, controlled benchmark of 16 distinct models spanning convolutional, hybrid, and transformer architectures to establish their efficacy in sparse volumetric anomaly detection under matched 2D and 3D configurations.
2) Empirical and mechanistic evidence demonstrating the conditions under which hierarchical windowed attention provides a systematic performance advantage over purely local (CNN) and purely global (ViT) inductive biases in medical volume analysis.
3) Mechanistic analysis using effective receptive field, attention-distance distributions, and centered kernel alignment to functionally link specific architectural properties to the resulting representational structure.
4) Generalizable design principles for model selection in medical imaging tasks involving sparse, multi-scale anomalies.

## II. RELATED WORK

### A. Vision Transformers and Hierarchical Attention

Vision Transformers [2] typically require far larger datasets than CNNs [13-15] because they lack built-in locality and equivariance, forcing them to learn spatial structure from data. Steiner et al. [16] showed that ViT requires 10-100× more training data than CNNs to reach equivalent performance without strong augmentation and regularization strategies. Hierarchical transformers (e.g., Swin, PVT) reintroduce locality and multi-scale structure through windowed or reduced-resolution attention, combining CNN-like spatial bias with transformer flexibility. The shifted window (Swin) Transformer [3, 17] restricts attention to local windows and alternates between standard and shifted window partitions across layers, achieving linear computational complexity while building multi-scale representations through progressive patch merging. Pyramid Vision Transformer (PVT) [5] applies spatial-reduction attention that lowers key–value resolution across layers. Both approaches reintroduce locality and scale hierarchy, key CNN inductive biases, while retaining learned attention rather than fixed convolutional kernels.

Empirical analysis reveals that hierarchical transformers learn hybrid representations. D'Ascoli et al. [18] showed that Swin's early layers exhibit CNN-like texture selectivity and locality, while deeper layers integrate global context through shifted windows. This progressive locality-to-globality differs fundamentally from CNNs, where receptive field growth is architecturally determined, and from standard ViTs, where all layers apply uniform global attention. The critical question for sparse anomaly detection is whether this intermediate strategy provides advantages when relevant features span intermediate spatial scales and training data are limited. Medical volumes differ substantially from natural images: they are isotropic or near-isotropic in three dimensions, exhibit sparse signal-to-noise characteristics in early pathology, and are often available in far smaller quantities. Prior work evaluating hierarchical transformers in medical imaging has primarily focused on dense prediction tasks with either massive pretraining or per-pixel supervision [19-21]. Their efficacy and inherent structural properties for tasks lacking dense supervision and defined by weak signals remains unexplored.

*B. Three-Dimensional Architectures for Medical Imaging*

Volumetric medical imaging has motivated diverse architectural strategies for handling three-dimensional data. Early 2D [22] and 2.5D [23] approaches discard cross-slice relationships or provide limited through-plane awareness without cubic memory requirements. On the other hand, 3D CNNs capture full volumetric context but rely heavily on convolutional locality. Native 3D architectures emerged with U-Net variants for segmentation. The 3D U-Net [24] and V-Net [25] established that joint volumetric feature extraction outperforms slice-based methods when computational resources permit, particularly for tasks requiring precise spatial localization. Recent work has incorporated attention mechanisms: UNETR [26] combines ViT encoders with CNN decoders, while Tang et al. [27] augment CNNs with self-attention blocks. However, these hybrid designs confound the contributions of convolution and attention, precluding controlled comparison of architectural inductive biases.

Prior 3D medical imaging work focuses overwhelmingly on segmentation (dense supervision) or classification of manifest diseases. Sparse anomaly detection requires integrating weak, non-adjacent signals without localization supervision. The architectural properties that benefit dense prediction (precise localization, boundary delineation) may differ from those optimal for integrating distributed weak signals. Our work provides the first systematic comparison isolating how pure CNN, pure transformer, and hierarchical transformer architectures handle sparse volumetric anomalies under matched training protocols.

*C. Architectural Inductive Biases and Feature Integration*

The relationship between architectural inductive biases and task characteristics determines model performance in data-limited regimes [28]. CNNs' locality constraints accelerate convergence for spatially local features but limit long-range integration. ViTs learn arbitrary spatial relationships but require massive data to discover structure CNNs encode architecturally [2, 15]. Hierarchical designs balance these extremes through progressive spatial integration. The relationship between receptive field scale and task characteristics has been studied primarily in natural-image settings. Luo et al. [29] showed that optimal receptive field size depends on object scale in detection tasks, with smaller fields benefiting fine-grained recognition and larger fields helping scene understanding. For medical anomaly detection, where subtle abnormalities may span a tiny fraction of the total voxel space, the question of receptive-field matching remain largely unexamined. Prior work reports architectural performance but offer limited examination of whether effective receptive field align with anomaly spatial extent.

*D. Volumetric Anomaly Detection and Open-Set Recognition*

Anomaly detection in volumetric data has primarily been addressed in industrial quality control [30] and video surveillance [31]. In medical imaging, anomaly detection overlaps with early disease detection and out-of-distribution identification. Reconstruction-based methods, including autoencoders and GANs, model normal data distributions and flag deviations but struggle with subtle changes and require careful threshold selection [32, 33]. Open-set recognition, which classifies known classes while rejecting unknown classes, shares conceptual similarities with anomaly detection but operates primarily on 2D data with categorical labels [34-36]. Recent discriminative approaches use metric learning to create compact feature clusters. Reciprocal point learning [37] explicitly models extra-class space by learning counterexamples, while orientational distribution learning [38] introduces hierarchical spatial attention for 2D open-set recognition. These methods demonstrate that attention to spatial arrangement, not merely feature distances, improves decision boundaries. Our work extends these principles to 3D medical imaging, showing that hierarchical attention provides similar benefits for sparse anomaly detection.

A fundamental limitation of prior anomaly-detection and open-set recognition work is reliance on categorical labels with hard boundaries between normal and abnormal. For progressive diseases with continuous progression, this framing introduces label noise where subclinical cases near decision boundaries receive inconsistent annotations across studies [39]. To address this, our approach eschews categorical prediction in favor of modeling disease severity as continuous risk score, derived from Gaussian mixture modeling of tomographic indices [11].

*E. Deep Learning for Subclinical Keratoconus Detection*

Deep learning methods have focused on 2D representations using CNN feature extractors [40-42], achieving strong performance on manifest KC but either did not evaluate subclinical cases or exhibited substantially reduced sensitivity in that regime. Only one prior study [43] has applied 3D CNNs to AS-OCT volumes for KC severity classification, reporting 95% accuracy without subclinical evaluation or architectural comparison. None have compared CNN and transformer families on volumetric corneal imaging. Importantly, subclinical detection is both the most clinically consequential stage and the point at which existing architectures show the largest performance gaps, suggesting that more expressive 3D inductive biases could meaningfully advance the field. Our approach therefore characterizes architectural behavior specifically in this early-disease regime.

## III. METHODOLOGY

The methodology is structured to facilitate a controlled architectural comparison of locality (CNNs), globality (ViTs) and hierarchy (Swin), for continuous KC prediction from three-dimensional anterior-segment optical coherence tomography (AS-OCT). Input data is prepared in both 2D slice and 3D volumetric configurations, which are then used to train 16 models. Performance is comprehensively assessed using regression, discrimination, and calibration metrics, and is complemented by architecture–performance analysis to functionally link architectural design choices to performance on sparse anomaly detection. The subsequent subsections detail the construction and execution of this unified experimental framework.

### A. Dataset and Continuous Risk Labels

The dataset comprises 12,579 AS-OCT volumes acquired using the MS-39 device (Costruzioni Strumenti Oftalmici, Florence, Italy) from 4,541 patients seen at Moorfields Eye Hospital (London, UK) between 2020 and 2024. Because this was an observational study using retrospective anonymized data collected in the course of routine clinical practice, explicitly reconsenting individual patients was not required as per UK Health Research Authority guidelines [44]. Each volume met strict image quality criteria: Placido coverage $\geq$ 65% (corneal surface visible in topography) or B-scan coverage $\geq$ 85% (scan lines with signal-to-noise ratio > 10dB). The AS-OCT volumes consist of 24 radial B-scans uniformly sampled across 360°, with a native resolution 1800×1024 pixels corresponding to 16×7 mm field of view with 9 μm axial resolution. Volumes were treated as Cartesian tensors for compatibility with standard 3D operations.

The soft label for an eye, $p_{KC} \epsilon [0,1]$, represents the posterior probability of KC. These labels were pre-computed and retrieved from prior validated work [11], where they were derived by applying Gaussian Mixture Modeling (GMM) to a vector of established tomographic indices $x$. The ground truth label is defined as:

$$p_{KC}(x) = \frac{\pi_2 N(x|\mu_2 \Sigma_2)}{\pi_1 N(x|\mu_1 \Sigma_1) + \pi_2 N(x|\mu_2 \Sigma_2)}, \quad (1)$$

where $N(\pi_i, \mu_{i,} \sum_i)$ denotes the multivariate Gaussian density function for component $i \epsilon \{1,2\}$ (Healthy and KC). From the 12,579 available volumes, we constructed a stratified subset of 1,456 volumes from 430 patients to ensure balanced representation across the probability continuum. Stratification was performed by dividing the [0,1] probability range into 10 equal bins and sampling proportionally from each bin, subject to the constraint that all volumes from a single patient remained together. The resulting distribution was:

- Healthy ($p_{KC} \leq 0.25$): 630 eyes
- SKC ($0.25 \leq p_{KC} \leq 0.75$): 220 eyes
- KC ($p_{KC} \geq 0.75$): 605 eyes

Five-fold cross-validation was performed, ensuring that all scans from a single patient were restricted to the same fold to prevent patient-level data leakage.

### B. Preprocessing and Input Configurations

All volumes were converted to float32 grayscale tensors and z-score normalized per-volume (raw AS-OCT intensities originally ranged from 0 to 255). Two input representations were established for controlled comparison of 2D slice-based vs. native 3D processing.

1) **2D Input Configuration.** A single representative B-scan was extracted from each volume at a fixed angular position of 90°, corresponding to the vertical meridian and highest likelihood of pathology location [45-47]. Each B-scan was bilinearly resampled to a canonical resolution of 224×224 pixels.
2) **3D Input Configuration.** The 24 radial B-scans were stacked along the angular dimension, yielding an initial tensor shape of (24, 1800, 1024). To standardize spatial dimensions and manage GPU memory, volumes were resampled to from native resolution (24×1800×1024) to isotropic voxel spacing of 143 μm using trilinear interpolation. The resample volumes (252×112×49 voxels) were then center cropped along the angular and width dimensions to yield fixed tensors of shape 112 × 112 × 80 voxels (16mm×16mm×11.4mm physical extent). Center cropping was performed by computing the spatial centroid and extracting a symmetric region of the target size.

Instance-level z-score normalization was applied to every input:

$$X' = \frac{X - \mu_x}{\sigma_x} \quad (2)$$

where $X$ is the input tensor, and $\mu_x$ and $\sigma_x$ are its mean and standard deviation (SD), respectively. Data augmentation included (1) random horizontal and vertical flips (p = 0.5), (2) random 3D rotations uniformly sampled from ±15°, and (3) elastic deformations generated by sampling random displacement vectors from a Gaussian distribution at a coarse grid (spacing = 10 voxels), smoothing with a Gaussian kernel (σ = 10 voxels), and scaling by magnitude factor α = 1.0 voxel. The displacement field was interpolated to full resolution and applied via cubic interpolation [48, 49]. Augmentations were applied on-the-fly.

### C. Architectural Families and Implementation

We compared 16 architectures categorized into four families based on their primary inductive bias (Fig. 2). All 2D models used standard ImageNet-1K pretrained weights. All 3D variants were trained from scratch due to the absence of large-scale pretrained 3D regression/classification models (mainly video and segmentation) or pretraining datasets:

1) **Convolutional Neural Networks:** We adopt ResNet-18 (11M parameters) and ResNet-50 (23M parameters) as canonical convolutional baselines due to their widespread use in medical imaging and their strong transfer performance under ImageNet-1k pretrained weights [1]. For 3D CNNs, we construct volumetric variants by replacing all 2D convolutions, pooling layers, and normalization layers with their 3D equivalents (Conv3d, MaxPool3d, BatchNorm3d), using 3×3×3 kernels and standard 3D residual blocks. The final feature map was reduced via global average pooling to a 512-dimensional embedding fed to a single-neuron regression head. We note that our 3D CNN baselines do not include recent large-kernel or attention-augmented architectures (e.g., ConvNeXt/ MedNeXt, non-local CNNs). Most available 3D variants of these families are video or segmentation-oriented encoder–decoders rather than established baselines for volumetric regression or classification, making direct comparison inappropriate.

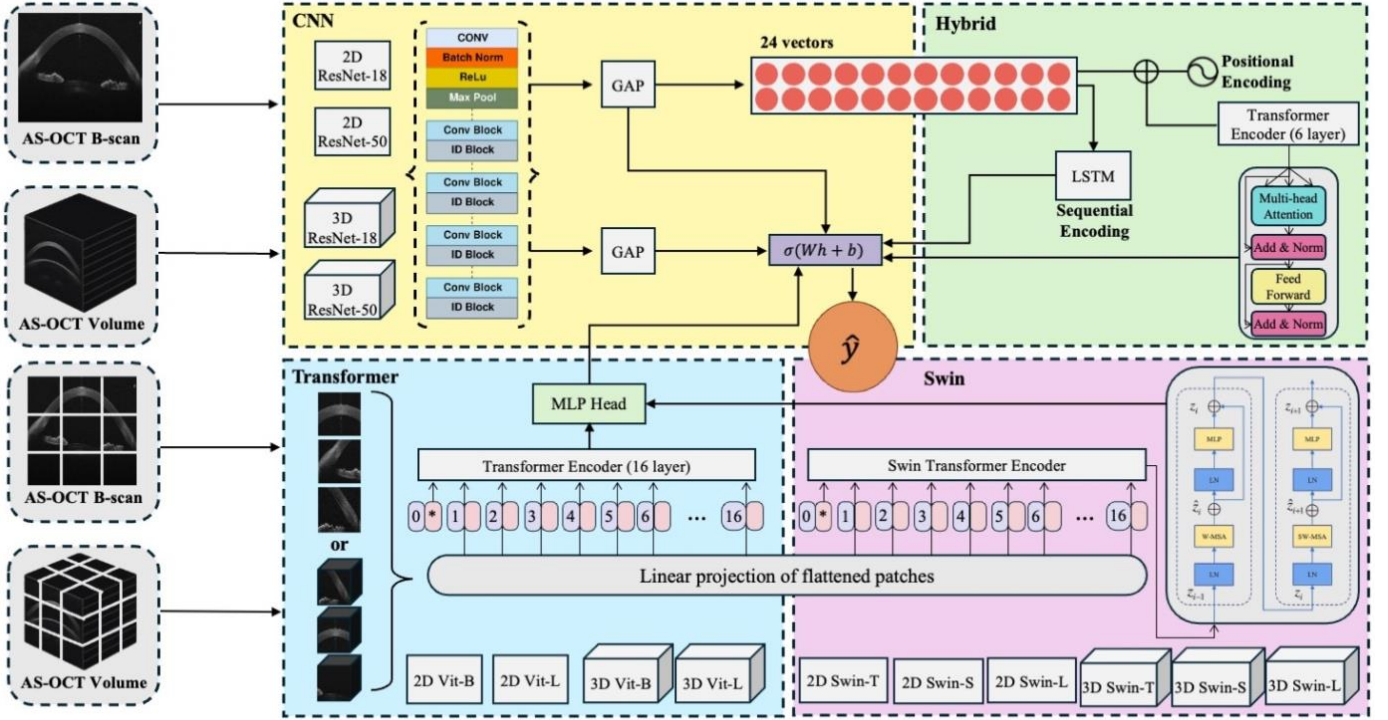

Fig. 2. Overview of the compared architectural families for subclinical keratoconus detection from AS-OCT B-scans and volumes. The study evaluates four inductive-bias families: (1) Convolutional Neural Networks (CNNs) implemented using 2D and 3D ResNet-18/50 backbones (yellow), where the 3D variants replace all convolutional, pooling, and normalization layers with Conv3d/MaxPool3d/BatchNorm3d and are trained from scratch; (2) Hybrid Sequential Models (green) in which a shared 2D ResNet-18 extracts 512-D feature vectors from each of the 24 constituent B-scans, followed by either a bidirectional LSTM (256 units) or a 6-layer Transformer encoder; (3) Vision Transformers (ViT) (blue), where inputs are partitioned into flattened 2D (16×16) or 3D (4×16×16) patches, linearly projected into tokens, augmented with positional embeddings, and processed by ViT-Base or ViT-Large encoders; and (4) Swin Transformers (purple), which employ hierarchical windowed self-attention and patch-merging in 2D (W=7×7 windows) and 3D (W=4×4×4) configurations, evaluated using Swin-Tiny/Small/Large. All models terminate in a regression head producing the predicted probability of keratoconus, $\hat{y}$. GAP: Global Average Pooling; MSA: Multi-head Self-Attention; W-MSA: Windowed Self-Attention; SW-MSA: Shifted-Window Self-Attention; LN: Layer Normalization; FFN: Feed-Forward Network; ID Block: Identity Block; MLP: Multi-Layer Perceptron.

2) **Hybrid Sequential Models:** These models decouple B-scan feature extraction from inter-slice context aggregation, mimicking the 2.5D approach. A shared 2D ResNet18 backbone was applied independently to each of the 24 B-scans to generate a sequence of vectors $Z = \{z_1, \ldots, z_{24}\}$, $z_i \in \mathbb{R}^{512}$. Two aggregation mechanisms are evaluated. In the first aggregator, the sequence $Z$ was processed by a bidirectional Long Short-Term Memory network with 256 units to represent recurrent models widely used in early medical volume analysis. The final hidden state vector was used for regression. In the second aggregator, the sequence $Z$ was augmented with a learnable [$CLS$] token and sinusoidal positional encodings, then processed by a 6-layer transformer encoder. Each layer consisted of multi-head self-attention (8 heads, 512-dimensional embeddings) followed by a feed-forward network (expansion ratio 4, yielding 2048 hidden dimensions). Layer normalization preceded each sub-block (pre-norm configuration), and dropout (p = 0.1) was applied after attention and feed-forward layers. The final [$CLS$] output was regressed to $p_{KC}$. Although these architectures offer stronger 2D multi-slice modeling than a single fixed-angle B-scan, they do not exhaust the space of competitive 2D baselines (e.g., multi-view omnidirectional slicing, learned slice selection, graph aggregation). These approaches may yield further gains but require new data pipelines and volumetric supervision beyond this study's scope.

3) **Vision Transformers:** In the 2D configuration, the 224×224 B-scan is decomposed into non-overlapping 16×16 2D patches or 4×16×16 3D patches, followed by standard transformer encoders. The input sequence $z_0$ was formed using linear projection (E), a learnable classification token ($x_{CLS}$) and positional embeddings (P):

$$z_0 = [x_{CLS}|Ex_1|Ex_2|\ldots|Ex_L] + P \quad (3)$$

We evaluate ViT-Base (D=768, 12 layers) and ViT-Large (D=1024, 24 layers) model scales, which remain the most stable ViT variants for medical imaging. We also note that several emerging efficient 3D transformer architectures (e.g., nnFormer, SegFormer3D) remain under active development and lack standardized pretraining pipelines; these are not included here.

4) **Swin Transformers (Swin-T/S/L):** In the 2D configuration, the model processing the single 224×224 B-scan. The architecture employed a 4-stage hierarchy with windowed self-attention confined to a local window size of W=7×7 and progressive patch merging. In the 3D configuration, the initial input was segmented in 4×4×4 cubic patches across all three dimensions. Attention was constrained locally to windows of size W=4×4×4, with multi-scale feature maps generated via alternating shifted windows and patch merging layers. Window sizes were selected to maintain comparable receptive field coverage relative to input dimensions: 7×7 for 224×224 2D inputs (3.1% linear coverage) versus 4×4×4 for 112×112×80 3D inputs (3.6% linear coverage in smallest dimension). We evaluate three model sizes: Swin-Tiny (29M params), Swin-Small (50M), and Swin-Large (197M). Although a Swin-B variant exists and is architecturally aligned with ViT-B in model size and compute, pretrained Swin-B weights are not consistently available across ophthalmic or general medical-imaging frameworks. Moreover, Swin-L offers a clearer high-capacity benchmark, while Swin-T and Swin-S span compact and mid-range configurations similar to our chosen ResNet and ViT models; therefore, Swin-B is excluded from the present study.

### D. Training Protocol and Optimization

All models were trained with AdamW optimizer ($\beta_1 = 0.9, \beta_2 = 0.999$) using cosine-decayed learning rates and early stopping based on validation mean squared error (MSE). Hyperparameters were tuned via random search, and the best checkpoint per fold was used for evaluation. A physical batch size of 16 was combined with gradient accumulation over 8 steps, yielding an effective batch size of 128. The optimization objective was the MSE between the predicted continuous probability $\hat{p}_{KC}$ and the soft label $p_{KC}$:

$$L_{MSE} = \frac{1}{N}\sum_{i=1}^{N}(\hat{p}_{KC}^{(i)} - p_{KC}^{(i)})^2 \qquad (4)$$

Training proceeded for maximum 50 epochs with early stopping based on validation MSE. Training was terminated if validation MSE did not improve by at least 0.001 for 3 consecutive epochs. The model checkpoint with the lowest validation MSE was retained for test evaluation. All models used PyTorch 1.13.1, CUDA 11.8, and NVIDIA RTX A6000 GPUs (48 GB VRAM).

### E. Evaluation Metrics

Model generalization and calibration were assessed using a comprehensive set of metrics on held-out test folds:
1) **Regression Metrics:** Mean squared error (MSE), mean absolute error (MAE), the coefficient of determination ($R^2$), and Pearson correlation coefficient ($\rho$).
2) **Discrimination Metrics:** Area under the receiver operating characteristic curve (AUROC) for the binary classification threshold $p_{KC} > 0.5$ (Healthy vs manifest KC).
3) **Calibration Metrics:** Brier score and assessment via reliability diagrams.

### F. Mechanistic Analysis

To investigate the architectural basis for performance differences, we performed three complementary analyses on the test set:

1) **Effective Receptive Field (ERF):** For each trained model, we estimated the ERF [29] by computing the absolute gradient of the predicted probability $\hat{y}$ with respect to the input volume **V**:

$$G = \left|\frac{\partial \hat{y}}{\partial \mathbf{V}}\right|, \qquad (5)$$

using a single backward pass from the regression output. Gradient maps were normalized to [0,1], and the ERF was defined as all voxels:

$$G(v) > 0.01 \cdot \max_{v} G(v). \qquad (6)$$

Layer-wise ERFs were obtained on the test set by backpropagating gradients to the input of each major architectural stage: the four residual blocks for CNNs, the slice-encoder and sequence-aggregator for hybrid models, each transformer encoder block for ViT-Base/Large, and each hierarchical stage (patch embedding through Stage 4) for Swin-Tiny/Small/Large. For every model, we quantified ERF size (voxel count above threshold), ERF radius (minimum bounding-sphere radius), and ERF growth profiles across depth.

2) **Attention Distance Distribution:** We quantified the attention-distance distribution for all transformer-based architectures, including ViT-B/L, Swin-T/S/L, and the transformer aggregator in the ResNet+ViT hybrid. For each attention block, we extracted the multi-head self-attention tensor $A \in \mathbb{R}^{H \times L \times L}$, where $H$ is the number of heads and $L$ the number of tokens. Token positions were mapped back to their corresponding spatial centroids in voxel coordinates. For each query token $I$, we identified the top-$k$ attended tokens ($k$=5), and for each attended pair ($i,j$) computed the Euclidean distance:

$$d_{ij} = \|p_i - p_j\|_2, \qquad (7)$$

where $p_i$ and $p_j$ denote the 3D coordinates of tokens $i$ and $j$. Distances were aggregated across all layers and heads, yielding per-model attention-distance distributions.

3) **Representational Similarity (CKA):** We employed centered kernel alignment to quantify the similarity of feature representations learned by different architectures and layers. For each model, we collected activation matrices from each major layer or block, flattened into $X \in \mathbb{R}^{N \times D}$, where $N$ is the number of test samples and $D$ the flattened feature dimension, and computed pairwise similarity with:

$$CKA(X,Y) = \frac{\|X^T Y\|_F^2}{\|X^T X\|_F \|Y^T Y\|_F}. \qquad (8)$$

We measured intra-model CKA (layer progression within the same architecture) and inter-model CKA (corresponding layers across different architectures, including CNN vs. ViT, ViT vs. Swin, Swin families, and hybrid vs. fully-3D encoders). Similarity matrices were averaged across folds.

*G. Auxiliary Tasks for Generalization Assessment*

To determine whether architectural trends generalize to other sparse signals beyond disease-related anomalies, two auxiliary tasks were trained as separate experiments. All 16 architectures were retrained from scratch on two balanced binary classification tasks using only healthy volumes:

1) **Age:** Binary classification of ≤30 vs. >30 years
2) **Sex:** Binary classification of male vs. female

These tasks were chosen because: (1) they involve entirely different biological patterns than keratoconus, testing architecture-level rather than feature-level learning; (2) they are independent of disease status, avoiding confounding; (3) ground truth is objective and unambiguous. Importantly, both tasks rely on subtle, spatially diffuse cues in AS-OCT volumes, making them a useful probe of whether architectural trends generalize beyond disease-specific morphology.

## IV. RESULTS

*A. Dataset Composition and Probability Distribution*

The architectural benchmarking cohort comprised 1,456 AS-OCT volumes obtained from 430 distinct patients. This dataset was systematically partitioned into training (n=301 patients, 957 volumes), validation (n=64 patients, 255 volumes), and test (n=65 patients, 244 volumes) sets, ensuring patient-level separation to prevent data leakage. The continuous disease risk label ($p_{KC}$) exhibited a trimodal distribution consistent with the underlying epidemiological states. The prevalence across the three risk categories was maintained across all partitions (training: 44%/15%/41%, validation: 43%/16%/41%, test: 42%/15%/ 43%, $\chi^2$=0.83, p=0.93) for consistency in training and evaluation.

Demographic analysis (Table 1) revealed expected epidemiological correlations: lower age was strongly associated with higher $p_{KC}$ values, with a mean age difference of approximately 7.5 years between the Healthy and KC groups (36.2±12.4 years vs. 28.6±9.7 years, p<0.001, Welch's t-test). Furthermore, the highest risk group ($p_{KC} \geq 0.75$) showed a greater proportional representation of Middle Eastern and South Asian ethnicities (54% combined) compared to the healthy cohort (38% combined, p<0.001, $\chi^2$ test), aligning with known population-level risk factors for KC.

*B. Predicting Keratoconus Probability from AS-OCT Imaging*

Table 2 presents the comprehensive performance metrics across all 16 architectures evaluated on the held-out test set (n=244 volumes, 65 patients). The final performance ranking generally segregated the models by both input dimension and architectural inductive bias: 3D hierarchical Swin > 2D Swin, 3D ViT and hybrid > 2D ViT > CNN. A performance advantage was observed for nearly all model utilizing 3D input over 2D counterparts. This improvement was evident across seven of the eight architecture pairs, with a modest but consistent mean increase of $\Delta_{AUROC}$=+0.054±0.015 (mean ± SD). Hierarchical Swin

TABLE 1
Dataset Demographics and Clinical Characteristics Stratified by Keratoconus Probability

| Characteristic | $p_{KC} \leq 0.25$ | $0.25 < p_{KC} < 0.75$ | $p_{KC} \geq 0.75$ |
|---|---|---|---|
| Patients (n) | 185 | 65 | 180 |
| Volumes (n) | 630 | 220 | 606 |
| Male (%) | 55 | 59 | 54 |
| Age (years, mean±SD) | 36.2 ± 12.4 | 30.8 ± 11.2 | 28.6 ± 9.7 |
| **Reported Ethnicity (%)** | 44 | | |
| White | 38 | 28 | 23 |
| Middle Eastern | 20 | 28 | 32 |
| South Asian | 18 | 21 | 22 |
| Black | 14 | 13 | 12 |
| East Asian | 7 | 7 | 6 |
| Mixed | 3 | 4 | 3 |

TABLE 2
Comprehensive Performance Comparison Across Architectural Families

| Model | Dim | P (M) | MSE↓ | MAE↓ | R²↑ | Pearson↑ | Brier↓ | AUROC↑ |
|---|---|---|---|---|---|---|---|---|
| **CNNs** | | | | | | | | |
| ResNet-18 | 2D | 11 | 0.051 | 0.17 | 0.59 | 0.75 | 0.21 | 0.72 |
| Resnet-50 | 2D | 24 | 0.049 | 0.16 | 0.62 | 0.78 | 0.20 | 0.74 |
| ResNet-18 | 3D | 33 | 0.047 | 0.16 | 0.68 | 0.83 | 0.19 | 0.81 |
| ResNet-50 | 3D | 46 | 0.044 | 0.15 | 0.72 | 0.85 | 0.18 | 0.82 |
| **Hybrids** | | | | | | | | |
| ResNet-18 + LTSM | 2D | 15 | 0.045 | 0.15 | 0.70 | 0.84 | 0.18 | 0.83 |
| ResNet-18 +ViT-B/16 | 2D | 101 | 0.043 | 0.15 | 0.73 | 0.85 | 0.17 | 0.85 |
| **ViT** | | | | | | | | |
| Vit-B/16 | 2D | 86 | 0.050 | 0.17 | 0.65 | 0.79 | 0.20 | 0.76 |
| Vit-L/16 | 2D | 307 | 0.048 | 0.16 | 0.66 | 0.80 | 0.19 | 0.77 |
| Vit-B/16 | 3D | 86 | 0.046 | 0.16 | 0.67 | 0.84 | 0.18 | 0.82 |
| Vit-L/16 | 3D | 307 | 0.045 | 0.16 | 0.67 | 0.85 | 0.18 | 0.82 |
| **Swin** | | | | | | | | |
| Swin-T | 2D | 29 | 0.047 | 0.16 | 0.68 | 0.83 | 0.19 | 0.80 |
| Swin-S | 2D | 50 | 0.045 | 0.15 | 0.71 | 0.85 | 0.18 | 0.83 |
| Swin-L | 2D | 197 | 0.044 | 0.15 | 0.71 | 0.86 | 0.18 | 0.83 |
| Swin-T | 3D | 29 | 0.042 | 0.15 | 0.75 | 0.87 | 0.17 | 0.85 |
| Swin-S | 3D | 50 | 0.037 | **0.14** | **0.79** | **0.89** | **0.15** | **0.89** |
| Swin-L | 3D | 197 | **0.036** | **0.14** | 0.78 | 0.88 | **0.15** | **0.89** |

*Dim: input dimension; P: Parameters (M = million). Vit: Vision Transformer; B: base; L: large; T: tiny; S: small; MSE: Mean Squared Error; MAE: Mean Absolute Error; R²: Coefficient of determination; Pearson: Pearson correlation coefficient; Brier: Brier Score; AUROC: Area Under the Receiver Operating Characteristic curve. All metrics computed on held-out test set (n=244 volumes). ↑ indicates higher is better, ↓ indicates lower is better. Bold indicates best performance. 95% bootstrap confidence intervals (10,000 iterations): AUROC ±0.03, R² ±0.04, MSE ±0.004.*

Transformers remained the most effective and parameter-efficient architectural family for this task. In 3D, Swin-S (50 parameters) and Swin-L (197 parameters) achieved the highest overall discrimination (AUROC=0.89) and lowest prediction error (MSE=0.036-0.037, Brier=0.15). Swin architectures demonstrated a clear superiority over global ViTs. The mid-sized 3D Swin-S outperformed the much larger 3D ViT-L (307M parameters, AUROC=0.82) by $\Delta_{AUROC}$=+0.07 while utilizing 84% fewer parameters. This efficiency advantage demonstrates that architectural inductive biases, not merely model capacity, determine performance in data-limited regimes.

The volumetric benefit was most substantial for CNN architectures ($\Delta_{AUROC}$=+0.08-0.09). However, a more pronounced effect was observed in the model's ability to explain prediction variance ($R^2$). While CNNs and Swin models showed appreciable $R^2$ increases in 3D ($\Delta R^2$=+0.09-0.10 and +0.07-0.08, respectively), pure 3D ViT models exhibited minimal $R^2$ improvement ($\Delta R^2$=+0.01-0.02). This suggests that although global attention leverages 3D input to better separate classes (AUROC improvement), it struggles to accurately model the continuous probability space, a core requirement of this regression task. Hybrid architectures performed competitively with the Swin family, achieving AUROC values of 0.83–0.85 despite using solely 2D input. Their performance overlapped with the 3D Swin-T and 2D Swin variants, indicating that the hybrid designs capture complementary spatiotemporal or local–global interactions that partially offset the absence of volumetric information. This places the hybrid models firmly within the middle-tier group of performers, alongside 3D ViTs, and 2D Swins.

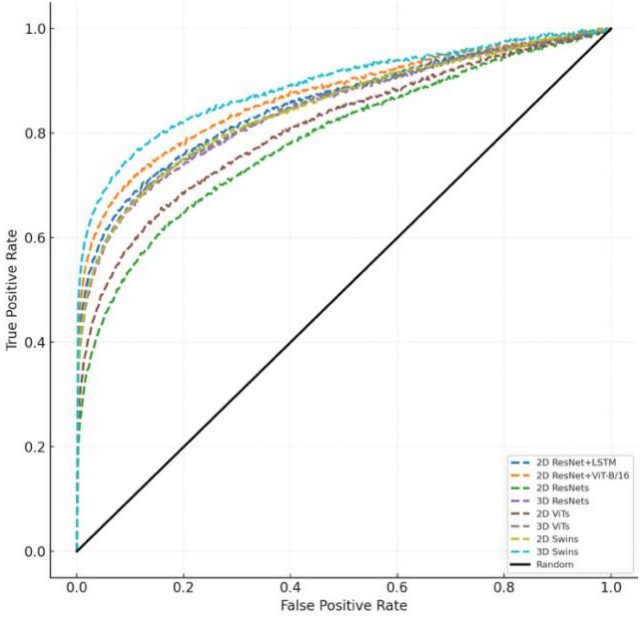

Fig. 3. Receiver operating characteristic curves comparing eight architectural categories across model families and input dimensionalities. Curves represent averaged performance for each category: CNN-2D, CNN-3D, Hybrid-2D, Hybrid-3D, ViT-2D, ViT-3D, Swin-2D, and Swin-3D. Three-dimensional variants outperform their 2D counterparts within each family, with Swin-3D achieving the strongest discrimination. The diagonal dashed line denotes random performance.

Within the Swin family, performance increased with model capacity, but the scaling trend was not strictly monotonic. From Tiny to Large, both 2D and 3D variants showed consistent improvement ($\Delta_{AUROC}$=+0.02-0.06, $p<0.01$, Delong's test). However, there appeared to be additional benefit scaling from Small to Large in 3D. Likewise, scaling ViTs from Base to Large in 3D produced no change in AUROC. Both observations suggest a limit in what the dataset can effectively support with increasing model size. The collective ROC curves in Fig. 3. illustrate a broad three-tier structure: the Swin 3D model occupy the top tier (AUROC=0.85-0.89), followed by the middle tier comprising 3D CNNs, hybrid models, 3D ViTs, and 2D

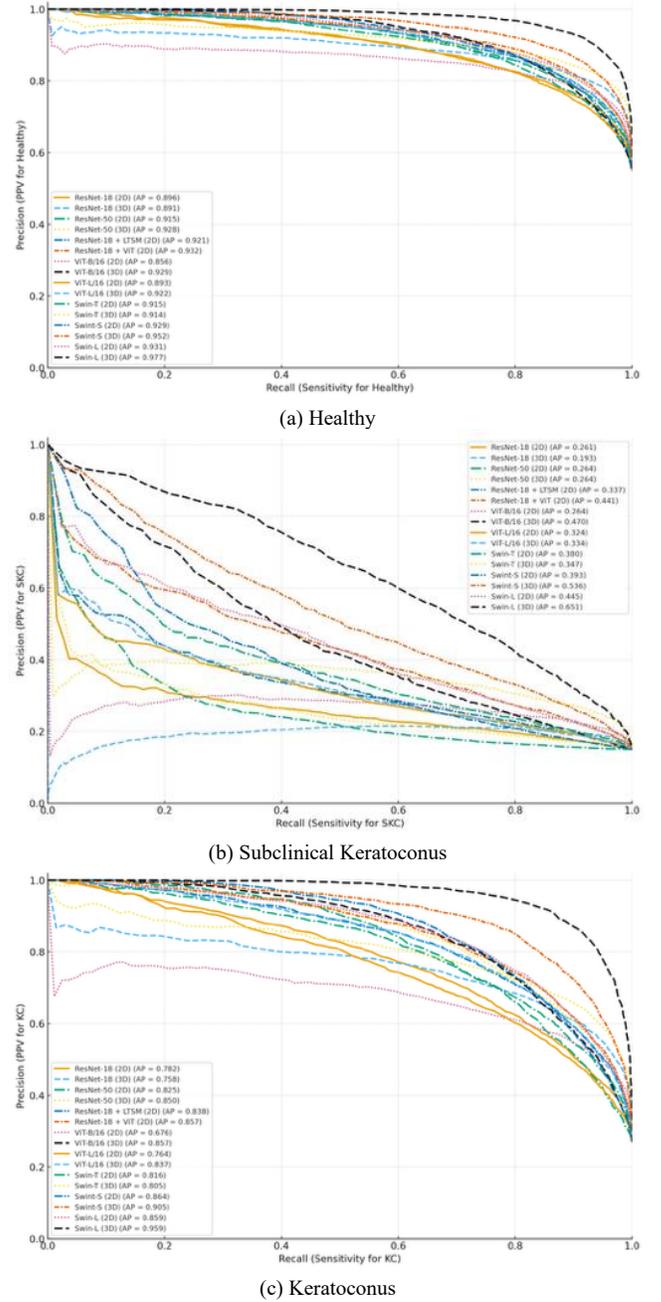

(a) Healthy

(b) Subclinical Keratoconus

(c) Keratoconus

Fig. 4. Precision-recall curve across three keratoconus probability bins.

Swins (AUROC=0.81-0.85), with 2D CNN/ViT architectures forming the lowest tier (AUROC 0.72-0.77).

*C. Sparse Anomaly Detection*

The influence of architectural design was most acutely observed in the differentiation of the mid probability (subclinical) range, where the anomaly signal is sparse and weak. Analysis of classification metrics (Table 3) demonstrated that performance differences were substantially amplified in this intermediate range compared to the low (Healthy) and high (KC) probability classes. Architectural capacity was largely compressed (Fig. 4) in the extreme ranges, with all models achieving relatively high performance (>0.80 sensitivity and specificity on healthy and

TABLE 3
Sensitivity and Specificity Stratified by Probability Range

| Model | Dim | Healthy Sens | Healthy Spec | Subclinical Sens | Subclinical Spec | Keratoconus Sens | Keratoconus Spec | Balanced Accuracy |
|---|---|---|---|---|---|---|---|---|
| CNNs | | | | | | | | |
| ResNet-18 | 2 | 0.80 | 0.81 | 0.64 | 0.58 | 0.82 | 0.80 | 0.76 |
| ResNet-50 | 2 | 0.81 | 0.83 | 0.57 | 0.65 | 0.84 | 0.82 | 0.78 |
| ResNet-18 | 3 | 0.82 | 0.83 | 0.58 | 0.63 | 0.85 | 0.83 | 0.78 |
| ResNet-50 | 3 | 0.83 | 0.84 | 0.59 | 0.69 | 0.86 | 0.84 | 0.80 |
| Hybrids | | | | | | | | |
| ResNet-18 + LTSM | 2 | 0.82 | 0.83 | 0.58 | 0.74 | 0.86 | 0.83 | 0.80 |
| ResNet-18 + ViT | 2 | 0.83 | 0.84 | 0.69 | 0.75 | 0.87 | 0.84 | 0.81 |
| ViT | | | | | | | | |
| Vit-B/16 | 2 | 0.80 | 0.81 | 0.61 | 0.69 | 0.83 | 0.81 | 0.77 |
| Vit-L/16 | 2 | 0.82 | 0.83 | 0.65 | 0.72 | 0.85 | 0.83 | 0.79 |
| Vit-B/16 | 3 | 0.82 | 0.83 | 0.67 | 0.73 | 0.85 | 0.83 | 0.79 |
| Vit-L/16 | 3 | 0.83 | 0.84 | 0.70 | 0.76 | 0.86 | 0.84 | 0.81 |
| Swin | | | | | | | | |
| Swin-T | 2 | 0.84 | 0.85 | 0.62 | 0.76 | 0.87 | 0.85 | 0.82 |
| Swin-S | 2 | 0.86 | 0.86 | 0.69 | 0.77 | 0.89 | 0.87 | 0.85 |
| Swin-L | 2 | 0.87 | 0.87 | 0.71 | 0.78 | 0.90 | 0.87 | 0.85 |
| Swin-T | 3 | 0.85 | 0.85 | 0.74 | 0.79 | 0.89 | 0.85 | 0.83 |
| Swin-S | 3 | 0.90 | 0.87 | 0.79 | 0.79 | 0.92 | 0.91 | 0.88 |
| Swin-L | 3 | 0.92 | 0.89 | 0.80 | 0.79 | 0.93 | 0.91 | 0.89 |

Healthy: $p_{KC} \leq 0.25$. Subclinical: $0.25 < p_{KC} < 0.75$. Keratoconus: $p_{KC} \geq 0.75$. Vit: Vision Transformer; B: base; T: tiny; S: small; L: large; Dim: dimensions; Sens: sensitivity; Spec: specificity. 95% confidence intervals (bootstrap, 10,000 iterations): Healthy/KC metrics ±0.02-0.03; Subclinical metrics ±0.10-0.14.

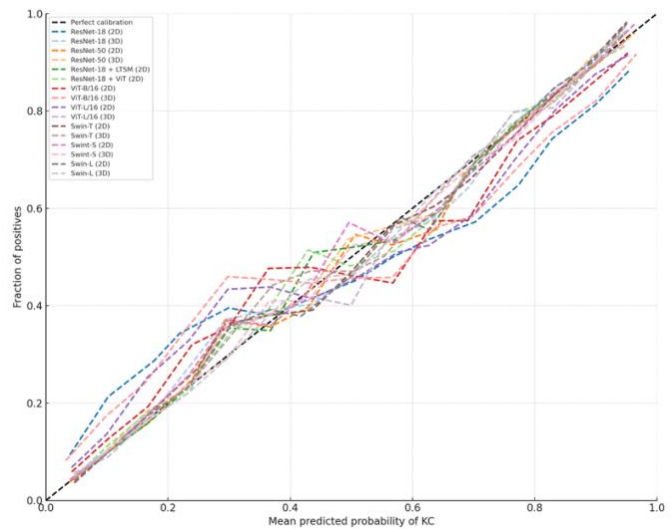

Fig. 5. Predicted vs. observed probability. Validation predictions were binned (equal-frequency deciles); for each bin, the x-axis shows the mean predicted probability of KC and the y-axis shows the empirical fraction of KC in that bin. The 45° dashed black line denotes perfect calibration. 3D Swin-L shows closest adherence to perfect calibration.

KC bins), indicating ceiling effects. In contrast, the subclinical group exhibited a stark performance spread. The maximum-to-minimum sensitivity range across all 16 architectures was 23 percentage points for the subclinical bin (0.80 for Swin-L 3D vs. 0.57 for ResNet-50). This variance was 2.0 times larger than the range observed for the Healthy and KC bins (12 percentage points each). The coefficient of variation for sensitivity was 0.12 in the subclinical regime, which is approximately 3 times higher than for Healthy (0.04) or KC (0.03), confirming that sparse anomaly detection is the discriminating regime for architectural comparison. However, the 95% bootstrap Cis for subclinical sensitivity were wider (±0.10-0.14), so these differences, while consistent across folds, should be interpreted as moderate rather than definitive.

Across architectures, 3D models and attention-based designs tended to overall improve subclinical detection, but the gains were neither uniform nor consistent. Several counter-examples were evident in Table 3. Some 3D CNNs achieved only marginal improvements or even slight degradations relative to their 2D baselines, and certain 2D Swin variants outperformed early 3D transformer models. These cases highlight that volumetric input alone does not guarantee better integration of subtle, cross-slice structure, and that architectural inductive biases play a larger role than dimensionality. Overall, the magnitude of improvement correlated with architectural sophistication. Although increased parameter count could partly explain some of these gains, the pattern was not monotonic with model size, and this concern is examined more formally in *Architectural Generalization*. Hybrid models and 3D ViTs achieved moderate gains over 2D CNNs, typically on the order of several percentage points, whereas hierarchical 3D Swin architectures produced the most pronounced increases. The strongest Swin model achieved an improvement of approximately 15-20 points over the 2D CNN baseline and a smaller advantage over the 3D ViTs, reflecting the benefits of combining hierarchical locality with volumetric processing. These findings indicate that subclinical detection benefits most from architectures that impose structured, multi-scale attention rather than from volumetric context or unconstrained global attention alone.

*D. Model Calibration*

Model calibration, the agreement between predicted probabilities and observed outcomes, is essential for clinical utility. It was assessed using reliability diagrams (Fig. 5.), constructed with equal-frequency decile binning on held-out test set predictions. Transformer-based models were generally better calibrated than CNNs. 3D Swin-S and Swin-L model showed the best calibration, with curve closely following the identity line. In contrast, the 3D Vit-L model showed moderate under-confidence within the 0.40-0.70 range (deviation up to 0.08), and the 3D ResNet-50 displayed systematic under-confidence across the 0.2-0.6 range (deviation up to 0.12). These trends suggest that hierarchical attention not only improves discrimination but also produces more reliable risk estimates than pure convolution or global attention under limited data.

*E. Mechanistic Analysis of Architectural Inductive Biases*

Effective receptive field analyses revealed systematic differences in how architecture utilize their theoretical spatial integration capacity (Table 4). CNNs exhibited underutilization despite deep networks, with ERFs saturating early and occupying only 24-31% of the available receptive

TABLE 4
Layer-wise Effective Receptive Field Radius

| Model | Dim | Kernel | Stage 1 | Stage 2 | Stage 3 | Stage 4 | E/T ratio |
|---|---|---|---|---|---|---|---|
| ResNet-18 | 2 | 3×3 | 5.1 | 9.8 | 10.3 | 12.5 | 0.24 |
| ResNet-50 | 2 | 3×3 | 5.4 | 9.1 | 10.9 | 13.2 | 0.26 |
| ResNet-18 | 3 | 3×3×3 | 6.1 | 9.9 | 11.5 | 14.1 | 0.31 |
| ResNet-50 | 3 | 3×3×3 | 6.2 | 9.5 | 10.9 | 13.4 | 0.28 |
| Model | Dim | Kernel | Encoder | | Aggregator | | E/T ratio |
| ResNet-18+LTSM | 2 | 3×3 | 12.5 | | 14.2 | | 0.52 |
| ResNet-18+ViT | 2 | 3×3 | 12.5 | | 16.9 | | 0.60 |
| Model | Dim | Patch | Stage 1 | Stage 2 | Stage 3 | Stage 4 | E/T ratio |
| Vit-B/16 | 2 | 16×16 | 6.9 | 9.8 | 12.1 | 12.8 | 0.88 |
| Vit-L/16 | 2 | 16×16 | 7.1 | 10.3 | 13.5 | 14.4 | 0.88 |
| Vit-B/16 | 3 | 4×16×16 | 7.3 | 11.4 | 14.3 | 17.1 | 0.88 |
| Vit-L/16 | 3 | 4×16×16 | 7.1 | 11.2 | 14.9 | 17.8 | 0.89 |
| Swin-T | 2 | 7×7 | 7.9 | 11.2 | 13.4 | 15.1 | 0.68 |
| Swin-S | 2 | 7×7 | 8.4 | 12.1 | 14.6 | 16.5 | 0.71 |
| Swin-L | 2 | 7×7 | 9.1 | 13.4 | 15.8 | 18.2 | 0.73 |
| Swin-T | 3 | 4×4×4 | 10.2 | 15.4 | 20.1 | 22.9 | 0.72 |
| Swin-S | 3 | 4×4×4 | 11.3 | 17.6 | 21.8 | 23.7 | 0.75 |
| Swin-L | 3 | 4×4×4 | 12.1 | 19.3 | 23.4 | 25.9 | 0.77 |

*Stages correspond to depth quartiles for ViT and hierarchical resolution stages for Swin models. Dim: input dimension; enc: encoder; agg: aggregator. E/T: effective/theoretical. Theoretical receptive field for ViT models is defined as the full token span. 2D models reported in pixels and 3D models reported in voxels.*

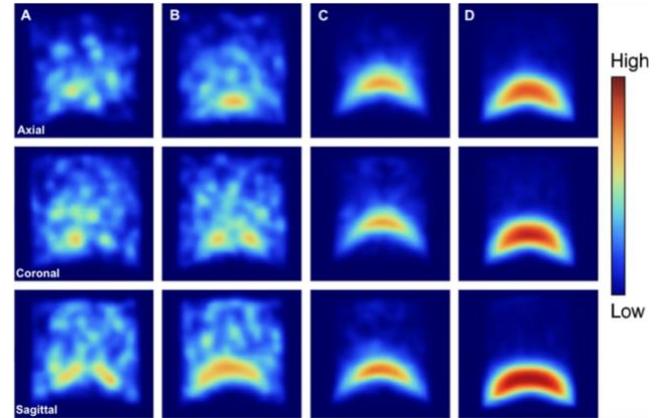

Fig. 6. Attention heatmaps for ViT and Swin models across axial, sagittal, and coronal views. ViT-B (A) and -L (B) exhibit diffuse, high-variance attention scattered throughout the volume, whereas Swin-S (C) and -L (D) show structured, localized attention concentrated around the posterior cornea and mid-stroma. Attention becomes increasingly focused with model capacity, most prominently in Swin-L. The Swin models emphasize the clinically significant region for early ectatic change, specifically the central cornea and stromal thickness axis, which aligns with the expected locus of subclinical keratoconus.

field. This early plateau reflects a well-known limitation of convolutional hierarchies: most gradient flow remains confined to local neighborhoods, preventing deeper layers from integrating broader spatial structure. Introducing attention into CNN encoders (e.g., ResNet-18+ViT) increased utilization efficiency substantially, although the absolute extent of integration remained below that of pure transformer-based models. ViTs, despite having theoretical access to the full token graph, did not fully exploit their global receptive fields. Their effective utilization remained moderate, and qualitative visualizations showed that gradients tended to cluster around localized regions rather than spreading uniformly across the volume. This indicates that global attention alone does not guarantee long-range integration; optimization dynamics tend to favor spatially constrained dependencies even when architectural constraints do not. Consistent with these observations, ViT attention maps (Fig. 6.) exhibited diffuse, high-variance patterns that frequently extended into peripheral, diagnostically irrelevant regions, suggesting inefficient allocation of attention capacity.

Hierarchical Swin Transformers achieved both the largest spatial integration and the highest utilization efficiency overall. Their receptive fields expanded progressively across stages, in contrast to the early saturation seen in convolutional networks. Utilization increased smoothly with model capacity, and 3D variants consistently integrated a broader spatial extent than their 2D counterparts. Among all families, Swin models approached the closest match between theoretical and realized spatial coverage, reflecting the effectiveness of shifted windows and multi-scale hierarchy for balancing local detail with mid-range global context. These findings nuance the local-versus-global dichotomy: neither convolutions nor unconstrained global attention alone make full use of their theoretical fields, whereas hierarchical attention provides a more balanced mechanism for leveraging available spatial context in volumetric imaging tasks.

*E. Attention Distance Distributions*

Attention-distance statistics demonstrated a clear, stage-dependent modulation of spatial integration demands (Table 5). Across all transformer-based architectures, subclinical presentations consistently elicited substantially longer attention distances than either healthy or overt KC cases. On average, subclinical volumes required 40–70% longer integration distances than healthy tissue and approximately 25–45% longer distances than overt KC. This pattern is biologically intuitive: healthy corneas exhibit broad, low-contrast structure that requires only moderate spatial aggregation, while KC produces sharply localised pathology that concentrates attention into compact, high-gradient regions. Subclinical disease, by contrast, presents with distributed, low-magnitude deviations, forcing models to integrate information across wider spatial neighborhoods to resolve ambiguous morphological cues.

Despite this overall pattern, the architectures differed in how they achieved such integration. Pure ViTs exhibited the longest and most variable attention distances, with subclinical cases producing increases frequently exceeding 50–70% relative to healthy volumes. These large jumps were accompanied by marked variance inflation, consistent with diffuse, anatomically inconsistent attention maps. This indicates that unconstrained global attention tends to scatter focus across both relevant and irrelevant regions, and that longer distances alone do not translate into improved performance. Hierarchical Swin Transformers, in contrast, exhibited moderately long yet considerably more stable

TABLE 5
Attention-Distance Statistics

| Model | Dim | Healthy | Subclinical | Keratoconus | Overall |
|---|---|---|---|---|---|
| ResNet-18+ViT | 2 | 9.3 ± 3.0 | 14.6 ± 4.9 | 11.4 ± 3.8 | 12.0 ± 4.1 |
| Vit-B/16 | 2 | 9.1 ± 3.4 | 15.2 ± 5.1 | 11.6 ± 4.0 | 12.0 ± 4.3 |
| Vit-L/16 | 2 | 9.4 ± 3.5 | 15.8 ± 5.3 | 12.0 ± 4.2 | 12.4 ± 4.5 |
| Vit-B/16 | 3 | 18.3 ± 5.2 | 32.1 ± 8.9 | 21.7 ± 6.8 | 23.1 ± 8.7 |
| Vit-L/16 | 3 | 18.2 ± 5.1 | 31.4 ± 8.7 | 21.3 ± 6.4 | 22.8 ± 8.4 |
| Swin-T | 2 | 10.4 ± 3.8 | 15.9 ± 5.1 | 12.5 ± 4.4 | 13.0 ± 4.6 |
| Swin-S | 2 | 11.6 ± 4.0 | 17.4 ± 5.4 | 13.4 ± 4.6 | 14.2 ± 4.8 |
| Swin-L | 2 | 12.1 ± 4.2 | 18.1 ± 5.8 | 14.0 ± 4.7 | 14.7 ± 5.0 |
| Swin-T | 3 | 16.2 ± 3.9 | 22.8 ± 5.3 | 18.6 ± 4.7 | 18.7 ± 5.2 |
| Swin-S | 3 | 16.5 ± 4.1 | 24.1 ± 5.9 | 18.9 ± 4.9 | 19.3 ± 5.6 |
| Swin-L | 3 | 16.8 ± 4.2 | 24.6 ± 6.3 | 19.1 ± 5.1 | 19.8 ± 5.8 |

*Dim: input dimension; MedDist: median attention distance (voxels); Pct>20: percentage of attention mass beyond 20 voxels, MaxDist: maximum Euclidean attended distance (voxels)*

attention distances. Across disease stages, Swin models expanded their integration distances by approximately 30–50% when moving from healthy to subclinical cases, then contracted by 20–30% when transitioning from subclinical to keratoconus. This adaptive "expand–contract" pattern reflects task-aligned modulation rather than indiscriminate global spread. Compared with ViTs, Swin models maintained both lower variance and higher anatomical coherence, consistently localizing attention to the posterior cornea and mid-stroma while still extending enough to capture the spatially distributed nature of subclinical abnormalities. This balanced behavior, neither overly local (CNNs) nor excessively global (ViTs), likely underpins the superior subclinical detection performance of hierarchical attention mechanisms.

Qualitative attention maps reinforced these quantitative trends (Fig. 6). CNN-based or sequence-augmented models remained sharply local; ViTs displayed scattered, high-variance attention with widespread peripheral excursions; and Swin models produced smooth, anatomically consistent crescents that respected corneal topology. The consistency of these patterns across disease stages suggests that hierarchical windowing provides a strong spatial inductive bias, enabling the model to expand integration by roughly one-third to one-half when subtle, distributed cues require it, while preserving anatomical structure and avoiding the overly diffuse behavior characteristic of pure global attention.

Centered kernel alignment analysis identified three clear representational clusters corresponding to the CNN, ViT, and Swin architectural families (Fig. 7). Within-family similarity was consistently high, while between-family similarity was substantially lower, indicating that architectural design strongly determines the structure of the learned feature space. Convolutional and vision transformer representations were the most distinct, reflecting the fundamental differences between localized convolutional filters and global attention mechanisms. Swin architectures formed an intermediate cluster positioned between CNN and ViT families. They shared moderate similarity with both groups, with a slight bias toward attention-based representations. This positioning is consistent with their hybrid inductive bias: early layers preserve localized structure extraction similar to CNNs, while deeper stages incorporate flexible long-range interactions characteristic of attention mechanisms. The Swin family also exhibited tighter clustering across model scales than either CNNs or ViTs, suggesting that hierarchical windowing produces more stable and reproducible representational structure across depth and parameter regimes.

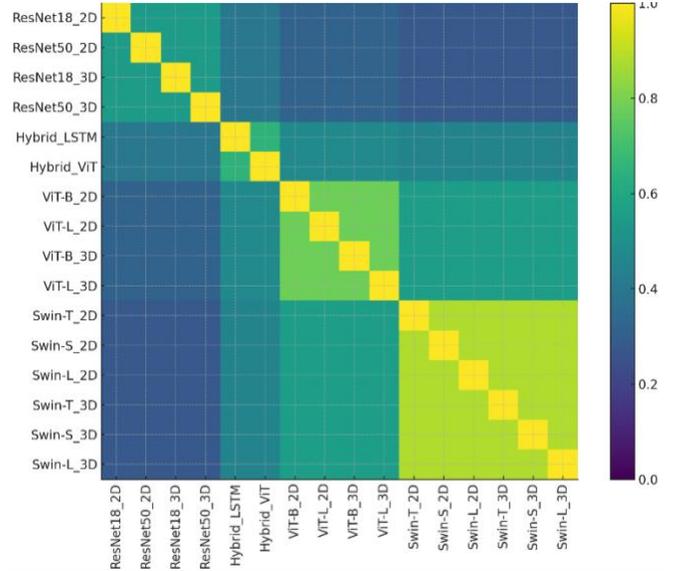

Fig. 7. Centered kernel alignment matrix between final-layer representations across all architectures, quantifying representational similarity from 0 (orthogonal) to 1 (identical up to linear transformation).

We did not quantify correlations between CKA structure and performance within the Swin family, because only three Swin variants were evaluated, which is insufficient for reliable statistical analysis. Instead, we interpret the CKA results descriptively. Swin models form a compact representational cluster whose members also occupy the upper range of detection performance, supporting the view that hierarchical attention learns a distinct and task-appropriate feature space for subtle volumetric anomalies. This alignment between representational coherence and functional behavior highlights the importance of structured inductive biases in 3D medical imaging tasks.

*F. Architectural Generalization*

The two auxiliary tasks probed different biological signals: age-related changes are subtle and spatially diffuse, whereas sex-related differences tend to produce stronger and more consistent patterns in corneal shape. Performance on age prediction was modest across architectures, consistent with the subtlety of the underlying signal, while sex classification achieved higher AUROCs. The overall architectural hierarchy resembled but did not replicate the ranking observed for keratoconus probability (Table 6). Hierarchical 3D Swin models generally occupied the top tier, with 3D ViTs and 3D CNNs forming a competitive middle tier. Although parameter count varies across these models, the observed reversals were not size-aligned: several larger

TABLE 6
Auxiliary Prediction Task Performance

| Model | Dim | Age (≤ 30 vs > 30) | | | | Sex (Male vs Female) | | | |
|---|---|---|---|---|---|---|---|---|---|
| | | Sens | Spec | Bal. Acc | AUROC | Sens | Spec | Bal. Acc | AUROC |
| ResNet-18 | 2 | 0.68 | 0.64 | 0.66 | 0.64 | 0.80 | 0.74 | 0.78 | 0.79 |
| ResNet-50 | 2 | 0.69 | 0.65 | 0.66 | 0.66 | 0.81 | 0.71 | 0.79 | 0.77 |
| ResNet-18 | 3 | 0.70 | 0.67 | 0.66 | 0.62 | 0.82 | 0.74 | 0.79 | 0.80 |
| ResNet-50 | 3 | 0.71 | 0.67 | 0.68 | 0.68 | 0.83 | 0.76 | 0.79 | 0.80 |
| ResNet-18+LSTM | 2 | 0.70 | 0.64 | 0.66 | 0.61 | 0.82 | 0.75 | 0.77 | 0.78 |
| ResNet-18+ViT | 2 | 0.72 | 0.65 | 0.69 | 0.65 | 0.82 | 0.75 | 0.80 | 0.82 |
| Vit-B/16 | 2 | 0.72 | 0.66 | 0.69 | 0.65 | 0.81 | 0.77 | 0.81 | 0.83 |
| Vit-L/16 | 2 | 0.71 | 0.66 | 0.70 | 0.69 | 0.80 | 0.75 | 0.80 | 0.83 |
| Vit-B/16 | 3 | 0.74 | 0.70 | 0.73 | 0.71 | 0.82 | 0.78 | 0.82 | 0.81 |
| Vit-L/16 | 3 | 0.72 | 0.69 | 0.72 | 0.73 | 0.83 | 0.80 | 0.80 | 0.85 |
| Swin-T | 2 | 0.73 | 0.68 | 0.71 | 0.73 | 0.81 | 0.80 | 0.85 | 0.85 |
| Swin-S | 2 | 0.75 | 0.69 | 0.72 | 0.73 | 0.84 | 0.80 | 0.83 | 0.90 |
| Swin-L | 2 | 0.75 | 0.70 | 0.75 | 0.74 | 0.84 | 0.83 | 0.86 | 0.88 |
| Swin-T | 3 | 0.77 | 0.72 | 0.76 | 0.78 | 0.85 | 0.81 | 0.87 | 0.90 |
| Swin-S | 3 | 0.78 | 0.72 | 0.76 | 0.78 | 0.87 | 0.83 | 0.85 | 0.90 |
| Swin-L | 3 | 0.78 | 0.73 | 0.76 | 0.79 | 0.90 | 0.84 | 0.87 | 0.91 |

*Dim: input dimensionality; Acc: accuracy; Sen: sensitivity; Spec: specificity; Bal. Acc.: balanced accuracy; Avg: average.*

models underperformed smaller ones, indicating that parameter budget alone does not explain these cross-task patterns. However, several informative deviations occurred. For example, on the age task a large 3D ViT slightly exceeded a smaller 3D Swin model, and some 2D Swin variants matched or surpassed certain 3D ViTs on sex classification. These reversals suggest that although hierarchical attention provides a broad advantage, it does not dominate uniformly across all forms of anatomical variation.

The performance gap between the strongest Swin model and the weakest CNN was smaller for these auxiliary tasks than for keratoconus probability. The AUROC difference between these extremes was typically in the range of 10 to 15 percent for age and slightly lower for sex. This indicates that while inductive bias continues to matter, the effect size is more modest when the task relies on global demographic characteristics rather than localized or subtly distributed pathology. To quantify cross-task consistency, we computed the Spearman rank correlation of AUROC across the three tasks: keratoconus probability, age, and sex. The resulting correlation was moderately high ($\rho = 0.78$, $p = 0.002$). This tempered value suggests that models sharing similar inductive biases tend to perform similarly across diverse tasks, yet task-specific cues, dataset size, and optimization dynamics interact meaningfully with architecture. Overall, these results support the interpretation that hierarchical attention confers a genuine but not universal advantage for subtle volumetric signals in the cornea.

## IV. LIMITATIONS AND FUTURE DIRECTIONS

This study reveals a fundamental principle for sparse anomaly detection: optimal architectural inductive bias must align with the spatial statistics of the pathological process itself. Diseases like KC do not begin as a focal lesion that gradually expands; rather, it emerges as a field effect where biomechanical instability manifests simultaneously across multiple stromal regions before coalescing into the characteristic apical cone. This explains why subclinical detection fundamentally differs from manifest disease classification: early pathology lacks a dominant focal signature that CNNs can latch onto, instead requiring integration of weak, correlated signals across anatomically distant but biomechanically coupled regions.

The hierarchical attention advantage therefore stems not from superior representational capacity per se, but from architectural resonance with disease phenomenology. Swin's progressive windowing mirrors the multi-scale nature of ectatic progression: cellular-level collagen disorganization (captured in early layers) correlates with regional thickness variations (middle layers) that collectively produce subtle global shape deformations (late layers). From the results, it can be suggested that CNNs fall short in this regime because they commit to locality too early. By the time deep layers theoretically access sufficient context, gradient flow has already concentrated on high-frequency texture, precluding integration of cross-region correlations. On the other hand, ViTs may fall short for the opposite reason: unconstrained global attention diffuses across the entire volume indiscriminately, treating anatomically distant, biomechanically uncoupled regions (e.g., central cornea vs peripheral limbus) as equally relevant, which dilutes signal from truly correlated structural deviations.

What remains unexplained is why optimization alone does not overcome these architectural constraints. Given sufficient data, shouldn't CNNs learn to propagate gradients more broadly, and shouldn't ViTs learn to concentrate attention appropriately? The persistent performance gaps despite hyperparameter tuning suggest that architectural priors shape the optimization landscape in ways beyond convergence speed - they fundamentally constrain which solutions are reachable within practical training regimes. If early disease detection inherently requires intermediate-scale spatial reasoning, then data-agnostic architectures (ViTs trained from scratch) will consistently underperform structure-aware designs (Swin) in medical imaging, regardless of dataset size.

### A. Rethinking the Volumetric Imperative

The volumetric advantage (mean $\Delta$AUROC=+0.05) is smaller than expected given that 3D models access 24× more spatial information. This challenges the assumption that "more data is always better" and suggests architectural capacity matters more than dimensional richness. The dissociation between ViT's discrimination improvement ($\Delta$AUROC=+0.06 in 3D) and failure to improve continuous modeling (mean $\Delta R^2$=+0.01) reveals that 3D ViTs separate healthy from KC cases using volumetric context but collapse continuous risk gradations into binary distinctions. This reflects how global attention aggregates information through averaging attended features across tokens, which is effective for binary classification but problematic for continuous risk estimation where spatial distribution of anomalies determines severity. Hierarchical attention preserves distinctions by maintaining multi-scale feature maps encoding both local severity (early layers) and global extent (late layers).

The success of 2D hybrid models (AUROC=0.83-0.85) further complicates this narrative. These architectures achieve competitive performance through explicit decomposition: ResNet-18 extracts per-slice features, then LSTM/Transformer aggregators model sequential dependencies. That this nearly matches 3D CNNs suggests architectural decomposition can substitute for volumetric convolution when inductive biases match data structure. However, hybrids underperform 3D Swin (ΔAUROC=-0.04-0.06), indicating learned sequential aggregation cannot fully replicate joint spatial-contextual reasoning. The gap likely reflects hybrids' inability to capture non-sequential cross-slice relationships between anatomically adjacent but sequentially distant regions.

*B. The Label Structure Problem and its Resolution*

Calibration failures in CNNs and ViTs (Brier=0.18-0.21 vs 0.15 for Swin) expose how these architectures learn from probabilistic labels. CNNs systematically under-predict risk in the 0.2–0.6 range because limited receptive fields capture only partial evidence. ViTs may overfit to strong discriminative features while ignoring subtle distributed cues, producing high confidence for clear cases but under-confidence for ambiguous presentations. Hierarchical attention explicitly models signal strength at multiple resolutions. Early layers capture high-frequency local details contributing weak evidence; middle layers aggregate regional patterns providing moderate evidence; late layers synthesize global morphology confirming or refuting assessments. This staged integration produces calibrated probabilities because predictions reflect accumulated evidence across scales rather than single dominant features. When scales agree, confidence is high; when scales conflict, uncertainty appropriately increases.

*C. The Generalizability Question: When Does Hierarchy Help?*

Auxiliary task results (age: ΔAUROC=8-12%, sex: ΔAUROC=12-15%) reveal hierarchical attention's advantage is task-dependent. Sex-related differences are spatially consistent and strong, properties CNNs can capture, explaining the smaller gap. Age-related changes are subtle and heterogeneous, requiring integration of diverse weak signals, better suited to hierarchical attention. Both tasks showed architectural reversals absent in KC prediction: some 3D ViTs matched smaller Swins on age, some 2D Swins exceeded 3D ViTs on sex. Hierarchical attention excels specifically for sparse, distributed, intermediate-scale anomalies. The "Goldilocks regime" – where pathology spans 5–30% of anatomical extent with weak per-region signatures but strong cross-region correlations – is where hierarchy provides decisive advantages.

This has practical implications: characterize anomaly spatial statistics (focal vs distributed, strong vs weak, scale) before choosing architectures. For punctate lesions, CNNs suffice; for organ-level pathology, global pooling or ViTs work; but for field effects like subclinical KC, hierarchical attention is necessary. The cross-task correlation ($\rho=0.78$) suggests these principles partially generalize, but task-specific validation remains essential.

*D. Theoretical Implications: Inductive Bias as Architectural Prior*

Performance gaps, particularly Swin's advantage despite comparable parameter counts, challenge the "bitter lesson" hypothesis that architectural inductive biases become irrelevant with sufficient data. Results suggest that for medical imaging, where datasets are limited by patient availability and annotation costs, inductive bias remains paramount. Hierarchical attention succeeds because its windowed, multi-scale structure encodes domain-general principles: (1) biological processes exhibit spatial locality, (2) pathology manifests at multiple scales simultaneously. These are general properties of anatomical organization, suggesting hierarchical architectures may constitute a broadly applicable prior for medical volume analysis. However, the study leaves open how design choices (window size, patch merging rate, depth) should be set for different anatomies. We used 4×4×4 windows for 112×112×80 volumes based on heuristics rather than principles. Future work deriving window configurations from anatomical correlation lengths could transform architecture design from empirical trial to theory-driven selection.

*D. Limitations and Future Directions*

Several mechanistic questions remain. Why does shifted-window attention specifically help versus standard windowing? Is it increased receptive field, preserved spatial locality, or gradient flow interactions? Why does hierarchical attention yield better calibration. Is it multi-scale evidence integration or implicit ensemble effects? Layer-wise probability calibration analysis could test these hypotheses.

Our architectural comparison isolated canonical representatives (ResNets for locality, ViTs for globality, Swin for hierarchy) for conceptual clarity but reduced ecological validity. The most significant omission is modern large-kernel CNNs (MedNeXt, ConvNeXt-3D), which achieve ViT-competitive performance through depthwise-separable convolutions with 7×7×7 or larger kernels. If large-kernel CNNs match Swin performance, the advantage may simply reflect that larger receptive fields help, regardless of mechanism. This matters theoretically (does integration mechanism matter?) and practically (large-kernel CNNs are simpler to deploy). We also omit efficient hierarchical transformers (SegFormer3D, Twins-3D) that reduce costs through spatial-reduction attention. These may close the Swin–ViT gap while matching ViT efficiency, weakening claims that hierarchy is necessary rather than sufficient. Our conclusions may overfit to tested architectures. ViT-L may underperform because standard ViT design is poorly aligned with multi-scale medical anomalies.

The asymmetry between 2D and 3D pretraining (2D models used ImageNet, 3D trained from scratch) tilted comparison against 3D superiority claims. The observed 3D advantage (ΔAUROC=+0.05) likely represents a lower bound. This reflects real-world constraints. Practitioners choose between mature 2D ecosystems and nascent 3D alternatives, evaluating practical utility rather than intrinsic capacity.

The omission of modern multi-view 2D approaches (DINOv2 feature extraction with learned aggregation) is relevant because hybrid models achieve strong performance (AUROC=0.83-0.85). If well-designed 2D pipelines match 3D Swin at lower computational cost, the practical recommendation shifts to lightweight 2D aggregation for screening and 3D for ambiguous cases. Future work should address this through expansive benchmarking. Our contribution lies in demonstrating that inductive bias choice substantially influences sparse anomaly detection—a conclusion that holds even if specific architectural rankings evolve.

Moreover, our implementation relies on Cartesian approximation of radial AS-OCT geometry, introducing distortion as neighboring voxels don't correspond to uniform angular distances. Implementing Swin in native cylindrical coordinates (r, θ, z) with cyclic angular wrapping would preserve spatial adjacency and may improve asymmetric pattern detection. Positional encodings would utilise trigonometric functions (sin/cos(kθ)) to represent rotational structure. This represents deeper domain knowledge integration: explicitly encoding radial symmetry into architecture rather than treating cornea as generic 3D volume.

Finally, the single-site, moderate-size dataset limits generalizability. External validation on independent cohorts with different devices (Pentacam, Anterion), demographics, and practice patterns is essential. Larger datasets may shift the performance landscape. Translating findings to practice requires addressing computational constraints. Swin-L's inference time and 45GB memory requirement exceed high-throughput screening constraints. Knowledge distillation or cascaded architectures (fast CNN pre-screening followed by Swin confirmation for ambiguous cases) may bridge this gap but require validation to ensure accuracy preservation during compression.

## V. Conclusion

Hierarchical attention provides consistent advantages for sparse volumetric anomaly detection, achieving strong performance and interpretable spatial reasoning. Through controlled comparison of 16 architectures for subclinical keratoconus detection, we demonstrated that 3D Swin Transformers achieve the highest performance (AUROC 0.89), parameter efficiency (5× fewer parameters than ViT-L for comparable performance), and clinical interpretability (attention aligned with pathological feature distributions). Its advantage stems from matching the spatial scale of subtle pathological features, avoiding the locality constraints of CNNs and the diffuse behavior of global attention. These findings extend to tasks characterized by subtle distributed signals and limited training data, providing practical guidance for model selection in medical volume analysis.

## Acknowledgment

L. Kandakji thanks Stephen Tuft, Dan Gore, and Shafi Balal for their clinical guidance.

## References

[1] K. He, X. Zhang, S. Ren, and J. Sun, "Deep residual learning for image recognition," in *Proceedings of the IEEE conference on computer vision and pattern recognition*, 2016, pp. 770-778.
[2] A. Dosovitskiy, "An image is worth 16x16 words: Transformers for image recognition at scale," *arXiv preprint arXiv:2010.11929,* 2020.
[3] Z. Liu *et al.*, "Swin transformer: Hierarchical vision transformer using shifted windows," in *Proceedings of the IEEE/CVF international conference on computer vision*, 2021, pp. 10012-10022.
[4] W. Wang *et al.*, "Pyramid vision transformer: A versatile backbone for dense prediction without convolutions," in *Proceedings of the IEEE/CVF international conference on computer vision*, 2021, pp. 568-578.
[5] T.-Y. Lin, P. Dollár, R. Girshick, K. He, B. Hariharan, and S. Belongie, "Feature pyramid networks for object detection," in *Proceedings of the IEEE conference on computer vision and pattern recognition*, 2017, pp. 2117-2125.
[6] G. Litjens *et al.*, "A survey on deep learning in medical image analysis," *Medical image analysis,* vol. 42, pp. 60-88, 2017.
[7] H.-J. Moon and S.-B. Cho, "A 4D Transformer with Spatiotemporal Attentions for Universal Diagnosis of Brain Disorders," *Neurocomputing,* p. 132068, 2025.
[8] H. Hashemi *et al.*, "The Prevalence and Risk Factors for Keratoconus: A Systematic Review and Meta-Analysis," *Cornea,* vol. 39, no. 2, pp. 263-270, Feb 2020, doi: 10.1097/ICO.0000000000002150.
[9] M. C. Arbelaez, F. Versaci, G. Vestri, P. Barboni, and G. Savini, "Use of a support vector machine for keratoconus and subclinical keratoconus detection by topographic and tomographic data," *Ophthalmology,* vol. 119, no. 11, pp. 2231-8, Nov 2012, doi: 10.1016/j.ophtha.2012.06.005.
[10] N. S. Bodmer *et al.*, "Deep Learning Models Used in the Diagnostic Workup of Keratoconus: A Systematic Review and Exploratory Meta-Analysis," *Cornea,* vol. 43, no. 7, pp. 916-931, 2024, doi: 10.1097/ico.0000000000003467.
[11] L. Kandakji *et al.*, "Data-driven detection of subclinical keratoconus via semi-supervised clustering of multi-dimensional corneal biomarkers," *Ophthalmology Science*.
[12] S. Kornblith, M. Norouzi, H. Lee, and G. Hinton, "Similarity of neural network representations

revisited," in *International conference on machine learning*, 2019: PMlR, pp. 3519-3529.

[13] M. Chen *et al.*, "Generative pretraining from pixels," in *International conference on machine learning*, 2020: PMLR, pp. 1691-1703.

[14] X. Chen, C.-J. Hsieh, and B. Gong, "When vision transformers outperform resnets without pre-training or strong data augmentations," *arXiv preprint arXiv:2106.01548,* 2021.

[15] M. Raghu, T. Unterthiner, S. Kornblith, C. Zhang, and A. Dosovitskiy, "Do vision transformers see like convolutional neural networks?," *Advances in neural information processing systems,* vol. 34, pp. 12116-12128, 2021.

[16] A. Steiner, A. Kolesnikov, X. Zhai, R. Wightman, J. Uszkoreit, and L. Beyer, "How to train your vit? data, augmentation, and regularization in vision transformers," *arXiv preprint arXiv:2106.10270,* 2021.

[17] X. Dong *et al.*, "Cswin transformer: A general vision transformer backbone with cross-shaped windows," in *Proceedings of the IEEE/CVF conference on computer vision and pattern recognition*, 2022, pp. 12124-12134.

[18] S. d'Ascoli, L. Sagun, G. Biroli, and A. Morcos, "Transformed CNNs: recasting pre-trained convolutional layers with self-attention," *arXiv preprint arXiv:2106.05795,* 2021.

[19] S. Mei, C. Song, M. Ma, and F. Xu, "Hyperspectral image classification using group-aware hierarchical transformer," *IEEE Transactions on Geoscience and Remote Sensing,* vol. 60, pp. 1-14, 2022.

[20] L. Wang *et al.*, "Multi-scale hierarchical transformer structure for 3d medical image segmentation," in *2021 IEEE International Conference on Bioinformatics and Biomedicine (BIBM)*, 2021: IEEE, pp. 1542-1545.

[21] A. He, K. Wang, T. Li, C. Du, S. Xia, and H. Fu, "H2former: An efficient hierarchical hybrid transformer for medical image segmentation," *IEEE Transactions on Medical Imaging,* vol. 42, no. 9, pp. 2763-2775, 2023.

[22] A. Prasoon, K. Petersen, C. Igel, F. Lauze, E. Dam, and M. Nielsen, "Deep feature learning for knee cartilage segmentation using a triplanar convolutional neural network," in *International conference on medical image computing and computer-assisted intervention*, 2013: Springer, pp. 246-253.

[23] H. R. Roth *et al.*, "A new 2.5 D representation for lymph node detection using random sets of deep convolutional neural network observations," in *International conference on medical image computing and computer-assisted intervention*, 2014: Springer, pp. 520-527.

[24] Ö. Çiçek, A. Abdulkadir, S. S. Lienkamp, T. Brox, and O. Ronneberger, "3D U-Net: learning dense volumetric segmentation from sparse annotation," in *International conference on medical image computing and computer-assisted intervention*, 2016: Springer, pp. 424-432.

[25] F. Milletari, N. Navab, and S.-A. Ahmadi, "V-net: Fully convolutional neural networks for volumetric medical image segmentation," in *2016 fourth international conference on 3D vision (3DV)*, 2016: Ieee, pp. 565-571.

[26] A. Hatamizadeh *et al.*, "Unetr: Transformers for 3d medical image segmentation," in *Proceedings of the IEEE/CVF winter conference on applications of computer vision*, 2022, pp. 574-584.

[27] H. Tang *et al.*, "Spatial context-aware self-attention model for multi-organ segmentation," in *Proceedings of the IEEE/CVF winter conference on applications of computer vision*, 2021, pp. 939-949.

[28] Z. Luo *et al.*, "Geodesc: Learning local descriptors by integrating geometry constraints," in *Proceedings of the European conference on computer vision (ECCV)*, 2018, pp. 168-183.

[29] W. Luo, Y. Li, R. Urtasun, and R. Zemel, "Understanding the effective receptive field in deep convolutional neural networks," *Advances in neural information processing systems,* vol. 29, 2016.

[30] P. Bergmann, M. Fauser, D. Sattlegger, and C. Steger, "Uninformed students: Student-teacher anomaly detection with discriminative latent embeddings," in *Proceedings of the IEEE/CVF conference on computer vision and pattern recognition*, 2020, pp. 4183-4192.

[31] M. Sabokrou, M. Fayyaz, M. Fathy, and R. Klette, "Deep-cascade: Cascading 3d deep neural networks for fast anomaly detection and localization in crowded scenes," *IEEE Transactions on Image Processing,* vol. 26, no. 4, pp. 1992-2004, 2017.

[32] T. Schlegl, P. Seeböck, S. M. Waldstein, G. Langs, and U. Schmidt-Erfurth, "f-AnoGAN: Fast unsupervised anomaly detection with generative adversarial networks," *Medical image analysis,* vol. 54, pp. 30-44, 2019.

[33] C. Baur, B. Wiestler, S. Albarqouni, and N. Navab, "Deep autoencoding models for unsupervised anomaly segmentation in brain MR images," in *International MICCAI brainlesion workshop*, 2018: Springer, pp. 161-169.

[34] A. Bendale and T. E. Boult, "Towards open set deep networks," in *Proceedings of the IEEE conference on computer vision and pattern recognition*, 2016, pp. 1563-1572.

[35] Z. Ge, S. Demyanov, Z. Chen, and R. Garnavi, "Generative openmax for multi-class open set classification," *arXiv preprint arXiv:1707.07418,* 2017.

[36] L. Neal, M. Olson, X. Fern, W.-K. Wong, and F. Li, "Open set learning with counterfactual images," in *Proceedings of the European conference on computer vision (ECCV)*, 2018, pp. 613-628.

[37] G. Chen, P. Peng, X. Wang, and Y. Tian, "Adversarial reciprocal points learning for open set recognition," *IEEE Transactions on Pattern Analysis and Machine Intelligence,* vol. 44, no. 11, pp. 8065-8081, 2021.

[38] Z.-g. Liu, Y.-m. Fu, Q. Pan, and Z.-w. Zhang, "Orientational distribution learning with hierarchical spatial attention for open set recognition," *IEEE Transactions on Pattern Analysis and Machine Intelligence,* vol. 45, no. 7, pp. 8757-8772, 2022.

[39] M. A. Henriquez, M. Hadid, and L. Izquierdo, Jr., "A Systematic Review of Subclinical Keratoconus and

Forme Fruste Keratoconus," *J Refract Surg,* vol. 36, no. 4, pp. 270-279, Apr 1 2020, doi: 10.3928/1081597X-20200212-03.

[40] A. Lavric, V. Popa, C. David, and C. C. Paval, "Keratoconus detection algorithm using convolutional neural networks: challenges," in *2019 11th International Conference on Electronics, Computers and Artificial Intelligence (ECAI)*, 2019: IEEE, pp. 1-4.

[41] K. Kamiya *et al.*, "Prediction of keratoconus progression using deep learning of anterior segment optical coherence tomography maps," *Annals of Translational Medicine,* vol. 9, no. 16, p. 1287, 2021.

[42] A. Elsawy *et al.*, "Multidisease deep learning neural network for the diagnosis of corneal diseases," *American journal of ophthalmology,* vol. 226, pp. 252-261, 2021.

[43] H. A. H. Mahmoud and H. A. Mengash, "Automated keratoconus detection by 3D corneal images reconstruction," *Sensors,* vol. 21, no. 7, p. 2326, 2021.

[44] G. U. "Information to be provided where personal data have not been obtained from the data subject. ." https://gdpr-info.eu/art-14-gdpr/ (accessed.

[45] L. Batres, D. Piñero, and G. Carracedo, "Correlation between anterior corneal elevation differences in main meridians and corneal astigmatism," *Eye & contact lens,* vol. 46, no. 2, pp. 99-104, 2020.

[46] J. C. Abad, R. S. Rubinfeld, M. Del Valle, M. W. Belin, and J. M. Kurstin, "Vertical D: a novel topographic pattern in some keratoconus suspects," *Ophthalmology,* vol. 114, no. 5, pp. 1020-1026. e4, 2007.

[47] D. A. Atchison *et al.*, "Peripheral ocular aberrations in mild and moderate keratoconus," *Investigative Ophthalmology & Visual Science,* vol. 51, no. 12, pp. 6850-6857, 2010.

[48] S. Olut, Z. Shen, Z. Xu, S. Gerber, and M. Niethammer, "Adversarial data augmentation via deformation statistics," in *European Conference on Computer Vision*, 2020: Springer, pp. 643-659.

[49] E. Castro, J. S. Cardoso, and J. C. Pereira, "Elastic deformations for data augmentation in breast cancer mass detection," in *2018 IEEE EMBS International Conference on Biomedical & Health Informatics (BHI)*, 2018: IEEE, pp. 230-234.

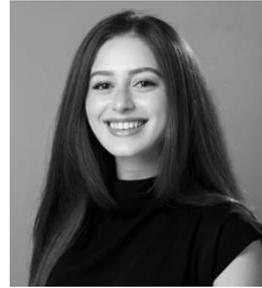

**Lynn Kandakji** received the bachelor's degree in medical technology and the master's degree in data science from the University of Ottawa, Ottawa, Canada. She received the PhD degree in computer science from University College London, London, UK. Her research interests include uncertainty modelling, pattern recognition, statistical learning for complex systems, and data-driven discovery.

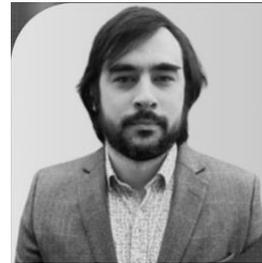

**William Woof** received the master's degree in mathematics from Durham's University, Durham, UK. He received the PhD degree in deep reinforcement learning with application to video-game AI from University of Manchester, Manchester, UK. His research interests include application of deep neural networks to unstructured representations. He has previously worked on projects using deep learning for video-game AI, identifying fuel-poor households, and applying deep reinforcement learning for simulating urban navigation.

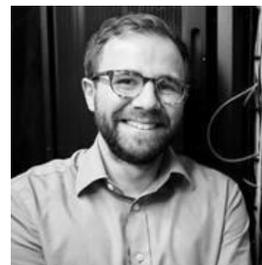

**Nikolas Pontikos** received the master's degree in computer science from University College London, London, UK and the master's degree in bioinformatics and theoretical systems biology from Imperial College London, London, UK. He received the PhD degree in statistical genetics and machine learning from Cambridge University, Cambridge, UK. His research interests include medical image analysis, ontologies, and genomic analysis.